\documentclass[sigconf]{acmart}
\makeatletter
\renewcommand\@formatdoi[1]{\ignorespaces}
\makeatother

\usepackage{todonotes}
\usepackage{graphicx}
\usepackage{multirow}
\usepackage{subcaption}
\usepackage{colortbl}

\AtBeginDocument{%
  \providecommand\BibTeX{{%
    \normalfont B\kern-0.5em{\scshape i\kern-0.25em b}\kern-0.8em\TeX}}}

\setcopyright{none}
\renewcommand\footnotetextcopyrightpermission[1]{} % removes footnote with conference information in first column
\begin{document}

%\title{When Recommendation Reveals Sensitive Information: Exploring Information Leakage from a Context-aware Recommender System}
% \title{There's Location in my List! Exploring User Attributes Implicit in Recommender System Output}
% \title[User Signals in Recommendation Lists]{User Signals in Recommendation Lists: Exploring How User Attributes in the Training Data Survive into Recommender Output}
% Title Suggestions
%\title{User Attributes in Recommender System Training Data and User Signals in Recommender Output}
% \title[Signal In Signal Out]{A Closer Look at User Attributes in Context-Aware Recommendation}
\title [Gender In Gender Out]{Gender In Gender Out: A Closer Look at User Attributes in Context-Aware Recommendation}

\author{Manel Slokom}
\email{m.slokom@tudelft.nl}
\affiliation{%
   \institution{Delft University of Technology}
  \country{Netherlands}
 }

\author{Özlem Özgöbek}
\email{ozlem.ozgobek@ntnu.no}
\affiliation{%
   \institution{Norwegian University of Science and Technology, Trondheim}
  \country{Norway}
 }

\author{Martha Larson}
\email{mlarson@science.ru.nl}
\affiliation{%
   \institution{Radboud University, Nijmegen}
  \country{Netherlands}
 }
\renewcommand{\shortauthors}{Slokom M., et al.}

\begin{abstract}
This paper studies user attributes in light of current concerns in the recommender system community: diversity, coverage, calibration, and data minimization.
In experiments with a conventional context-aware recommender system that leverages side information, we show that user attributes do not always improve recommendation. 
Then, we demonstrate that user attributes can negatively impact diversity and coverage.
Finally, we investigate the amount of information about users that ``survives'' from the training data into the recommendation lists produced by the recommender.
This information is a weak signal that could in the future be exploited for calibration or studied further as a privacy leak.
%Specifically, we study context-aware recommenders that add user side information to the training data.

\end{abstract}

\ccsdesc[500]{Information systems~Recommender systems}

\keywords{TopN recommendation, user attributes, context-aware recommendation}

\maketitle

\section{Introduction}
User attributes are included in the seminal data sets of recommender system research, e.g., MovieLens~\cite{Harper2015MovieLens}.
From the days of demographic recommender systems, mentioned in~\cite{burke2007hybrid}, attempts have been made to use user attributes to improve recommendation.
With the rise of Graph Neural Networks, interest in leveraging user attributes has been recently renewed~\cite{Huang2021Knowledge,do2022heterogeneous,Jiancan2022GCM}.
%add cite \cite{do2022heterogeneous} 
In this paper, we take a closer look at how helpful user attributes are in a conventional context-aware recommender system that makes use of user side information.
We study the impact of user attributes that go \emph{in}to a recommender system and the extent to which these attributes come \emph{out} of a recommender system, i.e., whether they strengthen the signal of user information that can be inferred from a user's recommendation list.
Our title mentions gender as a user attribute, but next to binary gender, we also investigate age, occupation and location.

We offer a broad, newly updated view on user attributes in recommendation, by reporting on a set of experiments that make three main contributions.
First, we demonstrate that user attributes are not always helpful to improve recommender prediction performance (Section~\ref{sec:recex}).
This point may not be surprising in light of the well-known disappointment of item attributes~\cite{pilaszy2009recommending}.
However, many papers studying side information combine item and user side information~\cite{dong2017hybrid, chen2018collective, zhou2021leverage}, rather than separating out user attributes as we do. 
Second, we show that user attributes actually have the potential to harm recommendation when we look beyond prediction performance to metrics like coverage and diversity (Section~\ref{sec:recdiv}).
Third, we study whether user attributes \emph{survive} from the training data into the recommender system output. 
We establish that there is a weak but consistent user signal in recommendation lists that can be detected by a machine learning classifier (Section~\ref{sec:recsuv}). 
Interestingly, adding user side information can amplify this signal without actually helping the recommender, opening new research questions for future work.

%Our paper makes the following contributions:
%\begin{itemize}
%\item User attributes do not always help to improve recommendation
%Sec 4 Recsys algo results
%\item User attributes diminish / reduce diversity and items coverage 
%Sec 5 Diversity and coverage
%\item User signals survive into the results
%Sec 6 Classification experiments
%\end{itemize}

\section{Related Work}
\label{sec:related}
In this section, we cover the related work that forms the background for each of our three contributions.

\subsection{Context-Aware Recommendation with User Side Information}
%Leveraging additional information such as contextual and side information to enhance the recommendations is a widely used practice. 
Context-aware recommenders integrate one or more of three types of side information: information related to users (e.g., age, gender), related to items (e.g., genre, price), and related to the interaction between users and items (e.g., time, location)~\cite{shi2014collaborative}.
In this paper, we focus on user side information because it is relatively less well studied than item side information and because of its potentially privacy sensitive nature, which makes it interesting and important to today's research community.

Use of user attributes in recommender systems dates at least back to demographic recommender systems~\cite{burke2007hybrid}, as previously mentioned.
Here, we briefly cover some examples of more recent collaborative filtering systems that have integrated user side information.
Variational Autoencoder approaches include~\cite{dong2017hybrid}, which stacks denoising auto-encoders (SDAE) to integrate side information into the latent factors, and~\cite{chen2018collective}, which uses a collective Variational Autoencoder (cVAE) for integrating side information for Top-N recommendation.
More recent work includes a clustering-based collaborative filtering algorithm that integrates user side information (such as age, gender and occupation) in a deep neural network~\cite{zhao2020clustering} and a Gaussian process based recommendation framework that leverages side information~\cite{zhou2021leverage}.
These approaches illustrate that researchers are interested in user attributes not just for improving cold start and sparsity, but also recommender performance across users.
For further examples of recent work, see~\cite{SI_comment2019, kulkarni2020context}.

In this work, we choose to focus on Factorization Machines (FMs)~\cite{rendle2010factorization}, classically used for context-aware recommendation.
FMs are a tried-and-true approach to context-aware recommendation, which allow easy integration of side information via extension of the user-item vector.
The advantage of FMs is that we can easily implement two recommender systems, one with and one without user attributes, and be confident that the use of the user attributes in the training data is the only difference between them.

\subsection{Diversity in Recommender Systems}
Diversity in recommender systems has drawn attention in recent years~\cite{Castells2015Diversity}.
Diversity can be defined as the potential of recommender system algorithms to recommend different or diverse content, e.g., recommending less popular items and targeting more niche items, while making personalized recommendation to users.
In~\cite{MatevZ2017Diversity}, the authors provide an overview of 
% research done on diversity in recommender system.
% They provided a summary of 
different definitions and measurements for diversity. 
Here, we are interested in the impact of side information on the diversification of the recommendation output.
Diversification is important for recommendations to be useful to the user.
Its importance is reflected in a surge of recent work on improving diversity, such as~\cite{Tommaso2017Adaptive}, a multi-attribute diversification method, and~\cite{steenvoorden2020attribute}, attribute-aware diversifying sequential recommender (SR).
Other work on diversity has focused on enhancing the user experience with system, such as~\cite{Tsai2017Leveraging}, which showed the importance of diversity.
In this paper, our focus is measuring diversity rather than attempting to improve it.

\subsection{User Signals in Recommender Output}
We are interested in whether recommendation lists contain a signal of user attributes and whether this signal is strengthened when the user attribute is explicitly part of the training data.
Previous work studying a user signal in recommender output is limited.
In~\cite{calandrino2011you}, the output of a recommender system is combined with a limited number of known transactions to infer unknown transactions of a target user. 
Our work is closer to~\cite{Beigi2020,zhang2021graph}, which focus on user attributes, specifically, infer  gender, age, and occupation of target users based on recommendation lists for those users combined with additional information.
In~\cite{Beigi2020}, the additional information is user embeddings that represent users internal to the recommender system.
In~\cite{zhang2021graph}, the additional information is the user's original profile, which is also internal to the recommender system.
To our knowledge, we are the first to carry out user attribute inference \emph{only} on the recommendations that were produced by the system without adding internal information.

Our interest in whether information in the training data is also present in the output of the recommender is reminescent of the idea of \emph{calibration}~\cite{Steck2018Calibrated}.
An uncalibrated recommender system has a mismatch between properties of the training data and of the output.
The properties conventionally studied in the literature, e.g., by~\cite{Steck2018Calibrated}, are item attributes.
Here we are looking at user attributes.
Like~\cite{Steck2018Calibrated} we find that consistency between the input and the output has an interesting impact above and beyond producing better recommendations in terms of prediction accuracy.

\section{Experimental Setup}
\label{sec:exp}
In this section, we first describe data sets.
Then, we describe the recommender system algorithms and classification algorithms that we will use in our experiments. 

\subsection{Data Sets}
Our experiments use three publicly available data sets.
First, we use two MovieLens data sets ML100K and ML1M~\cite{Harper2015MovieLens}.
We choose ML100K and ML1M because it includes demographic attributes of users such as gender, age, occupation, zipcode and also the timestamp needed for our temporal splitting.
We used zipcode to generate the State attribute. 
In order to convert MovieLens data from explicit feedback to implicit feedback, we set a \textit{cutoff}$ >= 3$, such that items with ratings $>=3$ are defined to be relevant, and the rest as non-relevant.  
Then, we pre-processed the resulting implicit data such that we have at least 20 interactions per user. 
ML100K subset contains 845 users and a total of 1574 movies for 80961 interactions.
ML1M subset contains a total of 5755 users and 3624 movies for 831745 interactions.
We also use a subset of LastFM~\cite{Thierry2011Million}, a music data set. 
We use artists as the items.
We preprocessed LastFM data, retaining only users who listened to at least 20 artists and artists to which at least 10 users have listened.
The result is a subset of 836 users and 12k artists.
For each user in LastFM data, gender and country location attributes are provided.
We used the Country attribute to generate the Continent and the EU vs Rest attributes.
We choose these data sets because they contain user attributes and they are publicly available. 
Table~\ref{tab:data} summarizes the statistics of the data sets.

\begin{table*}[!h]
\centering
\caption{Statistics of the data sets used for the experiments, including user attributes}
\label{tab:data}
% \resizebox{\columnwidth}{!}{%
\begin{tabular}{llllll}
\textbf{Data set}  & \textbf{\# Users}    &  \textbf{\# Items}  &  \textbf{\# Ratings}   &   \textbf{User attributes (\#Number of categories)}     \\ \hline

MovieLens (ML100K)  &  845 & 1574  & 80961 & Gender (2), Age (7), Occupation (21), State (52) \\ \hline
% Frappe   &  ... & ...  & ... & ... \% & Location, isWeekend \\ \hline
MovieLens (ML1M)   &  5755 & 3624  & 831745 & Gender (2), Age (7), Occupation (21), State (52) \\ \hline

LastFM subset   &  836 & 12155  & 501827 & Gender (2), Continent (7), EU vs Rest (2) \\ \hline
% \vspace{-0.5cm}
\end{tabular}%
% }
\end{table*}

\subsection{Recommender System Algorithms}
We generated our recommendation lists using Factorization Machines~\cite{rendle2010factorization} and also include BPRMF~\cite{Rendle2009BPR} for comparison. 
A Factorization Machine models pair-wise interactions with factorized parameterization and is suited to ranking problems with implicit feedback\footnote{For Factorization Machine implementation, we used RankFM toolkit: \url{https://rankfm.readthedocs.io/en/latest/}. Since RankFM does not include an implementation of hyper-parameter optimization, we implemented our own hyper-parameter optimization function by following initialization of hyper-parameters used in \url{https://github.com/lyst/lightfm}.}.
BPRMF is a matrix factorization algorithm using Bayesian personalized ranking for implicit data\footnote{For BPRMF implementation, we used Elliot Toolkit \url{https://elliot.readthedocs.io/en/latest/index.html}. We followed hyper-parameters optimization suggested in Elliot.}.
We used the RankFM implementation, including two variants of loss: Bayesian Personalized Ranking (BPR)~\cite{Rendle2014Pairwise} and Weighted Approximate-Rank Pairwise (WARP)~\cite{weston2013learning} to learn model weights via Stochastic Gradient Descent (SGD)~\cite{Rendle2014Pairwise}.
WARP loss is often described as performing 
better than BPR loss~\cite{abbas2019one,Maciej2015Metadata}. 
Our exploratory experiments confirmed that WARP loss was generally better than BPR loss, and we focus on WARP loss in our investigation.
User attributes (gender, age, occupation, and location) are one-hot-encoded before being used by FM. 
We note that in each run we add one user attribute at a time. 
In other words, we do not test combinations such as gender and location.
In this way, we can isolate the impact of the user attribute.

We used temporal splitting strategy such that we select 10\% of users' most recent interactions for test set, and 10\% of interactions for validation set and the 80\%, which are the remaining interactions, are used as training set.
We used validation set for tuning
hyper-parameters including: batch size, the learning rate (lr), user and bias regularization, and the number of latent factors.
For our factorization machine implementation \footnote{If accepted the code will be released on GitHub.}, 
we search for the best: lr in $\{0.001, .., 0.1\}$, number of training epochs in $\{5, .., 500\}$, and latent factor in $\{5, .., 200\}$. We left
alpha and beta parameters at default. 
Last but not least, we used MAP metric for optimizing hyper-parameters~\cite{shi2012TFMAP}.

In order to assess the quality of recommendation, we selected a set of commonly used TopN recommendation metrics namely, Precision (P@N), Recall (R@N), normalized Discounted Cumulative Gain (nDCG), and HR@N.
We set the size $ N $ of recommendation list to 50.
The diversity of recommendation lists is measured with item coverage, Gini index, and Shannon entropy~\cite{Castells2015Diversity}.
Item coverage computes the proportion of items that a recommender system can recommend from the entire items catalog.
Gini index and Shannon entropy are two different metrics used to measure the distributional inequality.
These measures take into account that an item is recommended to only some users, in addition to the item's distribution and to how many users it is recommended~\cite{Castells2015Diversity}.
The entropy based measure calculates the diversity offered by an item $i$ for a user $u$ in terms of the popularity of the item among the evaluated recommenders~\cite{Bellogin2010Heterogeneity}.

\subsection{Classification Algorithms}
We select two machine learning algorithms: Logistic Regression (LogReg) and Random Forest (RF) because they are widely used in literature~\cite{weinsberg2012blurme,chen2014effectiveness}.
We note that in our experimentation results we found that the LogReg classifier has close and comparable results to the RF classifier, with LogReg somewhat better.
In the remainder of the paper, for space reasons, we will focus on classification results from the LogReg classifier\footnote{Full tables of both classifiers including standard deviations can be found in the online folder \href{https://surfdrive.surf.nl/files/index.php/s/2wEglnYmGBQlgx4}{\nolinkurl{All_results.com}}.}.
We compare the performance of the Logistic Regression classifier to the performance of a random classifier using most frequent strategy (used as a baseline). 
Our classifiers take users' topN recommendation lists as the input.
We split our data using a stratified k-fold cross validation with $k=5$.
We measure the performance of classifiers using F1-score with macro-average. 
We choose F1-score because user attributes in our data sets are highly imbalanced.

\section{Leveraging User Attributes in the Recommender Input}
\label{sec:recex}
Our first experiment assesses the contribution that user side information makes to recommendation prediction performance when it is added to the training data.
Recommendation results from this experiment are shown in Table~\ref{tab:RecCARS}. 
We report topN recommendations measured with precision, recall, nDCG, and HR. 
We compare the performance of the Factorization Machine with WARP loss with and without (`None') side information.
We report BPRMF for comparison with FM without side information and confirm that the FM delivers better performance.
We note that results of FM using BPR loss are comparable, but not included here.

In Table~\ref{tab:RecCARS}, we observe it is possible to obtain improvements in recommendation performance when using user attributes as side information.
However, the improvements differ from one attribute type to another and from one data set to another.  
For ML100K and LastFM, recommendation with side information outperforms recommendation without side information.
For ML1M, we see that only the attribute \textit{State} helps to improve recommendation performance by a very small amount.
These results demonstrate that adding user attributes can possibly help, but is far from fail-safe strategy for improving recommendations.

\begin{table}[!h]
\centering
\caption{TopN (N=50) recommendation performance measured in terms of Precision@50 (P), Recall@50 (R), nDCG, and HR@50 on BPRMF and FM with WARP loss.}
\label{tab:RecCARS}
\resizebox{\columnwidth}{!}{%
\begin{tabular}{ccccccc}
\hline
\multirow{2}{*}{\textit{Data Sets}}       & \multirow{2}{*}{\textbf{Algorithms}} & \multirow{2}{*}{\textbf{\begin{tabular}[c]{@{}c@{}}User\\ Attributes\end{tabular}}} & \multicolumn{4}{c}{\textbf{Top-50 Recommendation}}    \\ \cline{4-7} 
                                          &                                      &                                                                                     & \textit{P} & \textit{R} & \textit{nDCG} & \textit{HR} \\ \hline
\multirow{6}{*}{\textbf{ML100K}} & \textit{BPRMF}                       & \textit{None}                                                                       & 0.2438     & 0.0383     & 0.0759        & 0.7479      \\ \cline{2-7} 
                                          & \multirow{5}{*}{\textit{WARP}}       & \textit{None}                                                                       & 0.2877     & 0.0444     & 0.0888        & 0.7751      \\ \cline{3-7} 
                                          &                                      & \textit{Gender}                                                                     & 0.3340     & 0.0522     & 0.1042        & 0.8462      \\ \cline{3-7} 
                                          &                                      & \textit{Age}                                                                        & 0.3210     & 0.0496     & 0.1002        & 0.8497      \\ \cline{3-7} 
                                          &                                      & \textit{Occupation}                                                                 & 0.3114     & 0.0493     & 0.0980        & 0.8414      \\ \cline{3-7} 
                                          &                                      & \textit{\begin{tabular}[c]{@{}c@{}}State\end{tabular}}                & 0.3268     & 0.0509     & 0.1015        & 0.8462      \\ \specialrule{.2em}{.1em}{.1em}
\multirow{6}{*}{\textbf{ML1M}}   & \textit{BPRMF}                       & \textit{None}                                                                       & 0.1519     & 0.0345     & 0.0582        & 0.6488      \\ \cline{2-7} 
                                          & \multirow{5}{*}{\textit{WARP}}       & \textit{None}                                                                       & 0.2135     & 0.0425     & 0.0743        & 0.7583      \\ \cline{3-7} 
                                          &                                      & \textit{Gender}                                                                     & 0.2028     & 0.0423     & 0.0731        & 0.7498      \\ \cline{3-7} 
                                          &                                      & \textit{Age}                                                                        & 0.1956     & 0.0415     & 0.0718        & 0.7359      \\ \cline{3-7} 
                                          &                                      & Occupation                                                                          & 0.1908     & 0.0417     & 0.0715        & 0.7225      \\ \cline{3-7} 
                                          &                                      & \textit{\begin{tabular}[c]{@{}c@{}}State\end{tabular}}                & 0.2216     & 0.0434     & 0.0768        & 0.7618      \\ \specialrule{.2em}{.1em}{.1em}
\multirow{6}{*}{\textbf{LastFM}} & \textit{BPRMF}                       & \textit{None}                                                                       & 0.2023     & 0.2209     & 0.2061        & 0.9474      \\ \cline{2-7} 
                                          & \multirow{5}{*}{\textit{WARP}}       & \textit{None}                                                                       & 0.2093     & 0.2035     & 0.1985        & 0.9665      \\ \cline{3-7} 
                                          &                                      & \textit{Gender}                                                                     & 0.2141     & 0.2178     & 0.2082        & 0.9677      \\ \cline{3-7} 
                                        %   &                                      & \textit{country}                                                                    & 0.2100     & 0.2122     & 0.2022        & 0.9605      \\ \cline{3-7} 
                                          &                                      & \textit{continent}                                                                  & 0.2101     & 0.2113     & 0.2015        & 0.9605      \\ \cline{3-7} 
                                          &                                      & \textit{EU vs Rest}                                                                      & 0.2154     & 0.2221     & 0.2103        & 0.9665      \\ \specialrule{.2em}{.1em}{.1em}
\end{tabular}%
}
\end{table}

\section{Diversity and Coverage of the Recommender Output}
\label{sec:recdiv}
Next, we move to investigate the impact of user side information on coverage and diversity.
Coverage is reported as the percent of items recommended and, diversity is measured in Shannon entropy and Gini index (higher is more diverse).
Table~\ref{tab:div_coverage} reports the results of recommendation using FM with WARP loss with and without user attributes.
We observe that compared to the recommender without side information (`None'), most user attributes depress coverage.
The exception is attributes with many values such as \textit{State} and \textit{Occupation} (in the case of ML100K).
We also observe that user attributes deteriorate diversity. (The exception is ML100K with \textit{Occupation} attribute.)
In some cases the drop is not very large, but the results support our conclusion that side information has the potential to harm recommendation.

\begin{table}[!h]
\centering
\caption{Item coverage and diversity of recommendation lists from Factorization Machine with WARP loss. }
\label{tab:div_coverage}
\resizebox{\columnwidth}{!}{%
\begin{tabular}{ccccc}
\hline
\textbf{Data Sets}               & \textit{User attributes} & \textit{\textbf{Items coverage}} & \textit{\textbf{Shannon Entropy}} & \textit{\textbf{Gini index}} \\ \hline
\multirow{5}{*}{\textbf{ML100K}} & \textit{None}             & 970                              & 8.9572                            & 0.2453                       \\ \cline{2-5} 
                                 & \textit{Gender}           & 825                              & 8.6667                            & 0.2005                       \\ \cline{2-5} 
                                 & \textit{Age}              & 867                              & 8.6623                            & 0.1994                       \\ \cline{2-5} 
                                 & \textit{Occupation}       & 1008                             & 8.9657                            & 0.2465                       \\ \cline{2-5} 
                                 & \textit{State}            & 1009                             & 8.8817                            & 0.2321                       \\ \specialrule{.2em}{.1em}{.1em}
\multirow{5}{*}{\textbf{ML1M}}   & \textit{None}             & 1988                             & 9.3369                            & 0.1361                       \\ \cline{2-5} 
                                 & \textit{Gender}           & 1703                             & 9.1161                            & 0.1159                       \\ \cline{2-5} 
                                 & \textit{Age}              & 1583                             & 8.8275                            & 0.0974                       \\ \cline{2-5} 
                                 & \textit{Occupation}       & 1606                             & 8.9066                            & 0.1017                       \\ \cline{2-5} 
                                 & \textit{State}            & 2002                             & 9.1889                            & 0.1249                       \\ \specialrule{.2em}{.1em}{.1em}
\multirow{4}{*}{\textbf{LastFM}} & \textit{None}             & 4508                             & 10.3876                           & 0.0969                       \\ \cline{2-5} 
                                 & \textit{Gender}           & 3866                             & 10.2768                           & 0.0858                       \\ \cline{2-5} 
                                 & \textit{Continent}        & 4010                             & 10.3793                           & 0.0918                       \\ \cline{2-5} 
                                 & \textit{EU vs Rest}       & 3044                             & 9.8399                            & 0.0622                       \\ \specialrule{.2em}{.1em}{.1em}
\end{tabular}
}
\end{table}
\section{User Signal in the Recommender Output}
\label{sec:recsuv}
Finally, we turn to explore the user signal in the recommendation lists.
First, we will discuss our classification results. 
Recall that we use a classifier to attempt to predict user attributes using the lists our recommender has output for each user.
We focus on Logistic Regression (LogReg) because it outperformed other classifiers we tested (in particular, Random Forest).
Results are shown in Table~\ref{tab:Classification}.
For comparison, we report scores of a random classifier with most frequent strategy as a baseline.
In all cases, our classifier outperforms this random baseline, which tells us that there is a user signal present in the recommender output.
Interestingly, both recommendation lists generated without user attributes (`None') and recommendation lists generated with user attributes contain at least a weak signal.

\begin{table*}[!h]
\centering
\caption{Classification results measured in terms of F1-score with macro-average. Recommendation lists are generated using FM with WARP loss. Random classifier uses most frequent strategy. The standard deviation over 5 folds is in between 0.000 and 0.0330.}
\label{tab:Classification}
\begin{tabular}{ccccccc} 
\hline
\textbf{Data Sets}               & \textit{\textbf{User Attributes}}                                & \textit{\textbf{Classification}} & \textit{Gender} & \textit{Age}       & \textit{Occupation} & \textit{State}     \\ 
\hline
\multirow{3}{*}{\textbf{ML100K}} & \multirow{2}{*}{None}                                            & \textit{Random}                  & 0.4198   & 0.1657     & 0.0741        & 0.0287       \\ 
\cline{3-7}
                                 &                                                                  & \textit{LogReg}                  & 0.4861±0.0330   & 0.2416      & 0.1103       & 0.0419      \\ 
\cline{2-7}
                                 & \begin{tabular}[c]{@{}c@{}}With side \\ information\end{tabular} & \textit{LogReg}                  & 0.5347   & 0.2219      & 0.1363       & 0.0579      \\ 
\specialrule{.2em}{.1em}{.1em}
\multirow{3}{*}{\textbf{ML1M}}   & \multirow{2}{*}{None}                                            & \textit{Random}                  & 0.4188   & 0.1820      & 0.0282       & 0.0569      \\ 
\cline{3-7}
                                 &                                                                  & \textit{LogReg}                  & 0.4958   & 0.2343      & 0.0891       & 0.0675      \\ 
\cline{2-7}
                                 & \begin{tabular}[c]{@{}c@{}}With Side\\ Information\end{tabular}  & \textit{LogReg}                  & 0.5022   & 0.2434      & 0.0866       & 0.0700      \\ 
\specialrule{.2em}{.1em}{.1em}
\multicolumn{3}{l}{}                                                                                                                   & \textit{Gender} & \textit{Continent} & \multicolumn{2}{c}{\textit{EU vs Rest}}  \\ 
\hline
\multirow{3}{*}{\textbf{LastFM}} & \multirow{2}{*}{None}                                            & \textit{Random}                  & 0.3666   & 0.3545      & \multicolumn{2}{c}{0.3416}        \\ 
\cline{3-7}
                                 &                                                                  & \textit{LogReg}                  & 0.4942   & 0.3912      & \multicolumn{2}{c}{0.5015}        \\ 
\cline{2-7}
                                 & \begin{tabular}[c]{@{}c@{}}With Side\\ Information\end{tabular}  & \textit{LogReg}                  & 0.5079   & 0.3917      & \multicolumn{2}{c}{0.5019}        \\
\specialrule{.2em}{.1em}{.1em}
\end{tabular}
\end{table*}

Next, to further understand this signal, in Table~\ref{tab:Classification}, we provide the raw difference and the percent change in classification performance on the recommendation lists before and after side information is added. (See rows labeled `Classification'.)
We interpret a relatively high percent change to mean that the information provided by a user attribute has \emph{survived} from the training data into the output data.
In cases, where the percent change is low, negative or zero, this information has become lost.
It is natural to expect that survival would depend on the type of user attributes or the number of values of a user attribute.
However, in Table~\ref{tab:Classification} we see that user attributes with the strongest survival vary across data sets.

Table~\ref{tab:Classification} also includes the Raw difference and Percent Change for recomendation results (nDCG).
We notice that survival is somewhat stronger across the board for ML100K and that this corresponds to a larger improvement in recommendation that are achieved by adding user attributes.
However, overall there is no indication of a clear and simple relationship of the usefulness of user attributes to a recommender system and the survival of those attributes in recommender system output.
We discuss the implication of this finding the final section of the paper.

\begin{table*}[!h]
\centering
\caption{Survival signal and recommendation improvement reported as absolute difference and relative change in classification between recommendation without user side information and recommendation with user side information. Recommendation lists are generated using FM with WARP loss and values calculated using nDCG metric. The negative values in $\% Change$ of recommendation mean that user attribute does not help to improve recommendation performance, but made it worse. The negative values in $\% Change$ of classification mean that user signal did not survive to the recommendation outputs. }
% Raw difference =$ F1_{attribute}^{LogReg} (With\_side\_information) - F1_{attribute}^{LogReg} (None) $ and $ nDCG_{attribute}^{WARP} (With\_side\_information) - nDCG{attribute}^{WARP} (None) $
\label{tab:RelativeChange}

\begin{tabular}{ccccccc}
\hline
\textbf{Data Sets}               & \multicolumn{2}{c}{\textit{Task}}                                  & \textit{Gender}                     & \textit{Age}                           & \textit{Occ}      & \textit{State}      \\ \hline
\multirow{4}{*}{\textbf{ML100K}} & \multirow{2}{*}{Classification} & \textit{Raw difference} & 0.0486                              & -0.0197                                & 0.026             & 0.0160              \\ \cline{3-7} 
                                 &                                          & \textit{\% Change}      & 0.1000                                 & -0.0800                                & 0.2400            & 0.3800              \\ \cline{2-7} 
                                 & \multirow{2}{*}{Recommendation}          & \textit{Raw difference} & 0.0154                              & 0.0114                                 & 0.0092            & 0.0127              \\ \cline{3-7} 
                                 &                                          & \textit{\% Change}      & 0.1700                              & 0.1300                                 & 0.1000            & 0.1400              \\ \specialrule{.2em}{.1em}{.1em}
\multirow{4}{*}{\textbf{ML1M}}   & \multirow{2}{*}{Classification}          & \textit{Raw difference} & 0.0064                              & 0.0091                                 & -0.0025           & 0.0025              \\ \cline{3-7} 
                                 &                                          & \textit{\% Change}      & 0.0100                              & 0.0390                                 & -0.0300           & 0.0370              \\ \cline{2-7} 
                                 & \multirow{2}{*}{Recommendation}          & \textit{Raw difference} & -0.0012                             & -0.0025                                & -0.0028           & 0.0025              \\ \cline{3-7} 
                                 &                                          & \textit{\% Change}      & -0.0200                             & -0.0300                                & -0.0400           & 0.0300              \\ \specialrule{.2em}{.1em}{.1em}
\multicolumn{3}{l}{}                                                                                  & \multicolumn{1}{l}{\textit{Gender}} & \multicolumn{1}{l}{\textit{Continent}} & \multicolumn{2}{l}{\textit{EU vs Rest}} \\ \hline
\multirow{4}{*}{\textbf{LastFM}} & \multirow{2}{*}{Classification}          & \textit{Raw difference} & 0.0137                              & 0.0004                                 & \multicolumn{2}{c}{0.0005}              \\ \cline{3-7} 
                                 &                                          & \textit{\% Change}      & 0.0300                              & 0.0000                                  & \multicolumn{2}{c}{0.0000}               \\ \cline{2-7} 
                                 & \multirow{2}{*}{Recommendation}          & \textit{Raw difference} & 0.0097                              & 0.0118                                 & \multicolumn{2}{c}{0.0030}              \\ \cline{3-7} 
                                 &                                          & \textit{\% Change}      & 0.0500                              & 0.0600                                 & \multicolumn{2}{c}{0.0200}              \\ \specialrule{.2em}{.1em}{.1em}
\end{tabular}
\end{table*}
\section{Conclusion and Future Work}
In this paper, we have studied a conventional Factorization Machine over which we exercise tight control.
We have shown that user attributes do not always help recommendation and can harm coverage and diversity.
Our results point to the need for caution with user attributes.
The survival of user information into recommender output constitutes a privacy leak of the sort that has concerned~\cite{Beigi2020,zhang2021graph}, but here measured without access to recommender-internal information.
Future work must avoid increasing the user signal in the recommendation list of a user without good cause in order to protect user privacy and respect data minimization. 

Future work should also extend the possible parallel with calibration~\cite{Steck2018Calibrated}. 
More research is necessary to gain insight into how measuring or manipulating the match between user attributes in the input and the output can be used to understand and improve recommender systems, also moving beyond accuracy.

\bibliographystyle{ACM-Reference-Format}
\bibliography{LateBreakingResults}


\end{document}